%% file: main.tex
\documentclass[10pt,twocolumn,letterpaper]{article}

\usepackage{cvpr}              

\input{preamble}

\usepackage[accsupp]{axessibility}
\usepackage{array}
\definecolor{lightgray}{gray}{0.9}
\definecolor{cvprblue}{rgb}{0.21,0.49,0.74}
\usepackage[pagebackref,breaklinks,colorlinks,citecolor=cvprblue]{hyperref}

\usepackage{colortbl}
\usepackage{xcolor}
\definecolor{lightgray}{gray}{0.9}
\usepackage{booktabs}
\usepackage{color, soul}

\definecolor{lightgray}{gray}{0.9}
\definecolor{lightblue}{rgb}{0.93,0.95,1.0}
\definecolor{darkgreen}{rgb}{0.0,0.6,0.0}
\definecolor{darkblue}{rgb}{0.0,0.0,0.5}
\definecolor{pinegreen}{rgb}{0.0, 0.47, 0.44}
\definecolor{deepmagenta}{rgb}{0.8, 0.0, 0.8}
\definecolor{amber}{rgb}{1.0, 0.49, 0.0}

\newcommand{\ignorebig}[1]{}

\def\Secref#1{Section~\ref{#1}}

\newcommand{\minisection}[1]{\noindent{\textbf{#1}.}}
\newcommand{\tabref}[1]{Table~\ref{#1}}

\newcommand{\figgref}[1]{Figure~\ref{#1}}

\newlength\savewidth

\newcommand{\model}{Compositional Chain-of-Thought}
\newcommand{\smodel}{CCoT}
\newcommand{\gcol}[1]{{\bf \fontsize{6.5}{42}\selectfont \color{citecolor!80}~(#1)}}

\definecolor{citecolor}{RGB}{34,139,34}
\definecolor{lightred}{RGB}{241,140,142}
\definecolor{amber(sae/ece)}{rgb}{1.0, 0.49, 0.0}
\definecolor{battleshipgrey}{rgb}{0.52, 0.52, 0.51}
\definecolor{cadmiumorange}{rgb}{0.93, 0.53, 0.18}
\definecolor{applegreen}{rgb}{0.55, 0.71, 0.0}
\definecolor{cadmiumgreen}{rgb}{0.0, 0.42, 0.24}
\definecolor{forestgreen}{rgb}{0.13, 0.55, 0.13}
\definecolor{red}{rgb}{0.89, 0.0, 0.13}

\def\paperID{16843} 
\def\confName{CVPR}
\def\confYear{2024}

\title{Compositional Chain-of-Thought Prompting for Large Multimodal Models}

\author{Chancharik Mitra
\quad Brandon Huang 
\quad Trevor Darrell 
\quad Roei Herzig
\\
University of California, Berkeley 
}

\begin{document}
\maketitle
\input{sec/0_abstract}    
\input{sec/1_intro}
\input{sec/2_related_works}
\input{sec/3_methods}

\input{sec/4_evaluation}

\input{sec/5_results}

\input{sec/6_conclusion}
\input{sec/7_impacts}
\subsubsection*{Acknowledgements.}
\vspace{-5pt}
We would like to thank Suzie Petryk, Alon Mendelson, Sanjay Subramanian, Rudy Corona, and Leonid Karlinsky for helpful feedback and discussions. This project was supported in part by DoD, including PTG and/or LwLL programs, as well as BAIR's industrial alliance programs.

{
    \small
    \bibliographystyle{ieeenat_fullname}
    \bibliography{main}
}


\appendix
\input{sec/supp}

\end{document}

%% file: preamble.tex
\usepackage[dvipsnames]{xcolor}


%% file: sec/0_abstract.tex
\begin{abstract}

The combination of strong visual backbones and Large Language Model (LLM) reasoning has led to Large Multimodal Models (LMMs) becoming the current standard for a wide range of vision and language (VL) tasks. However, recent research has shown that even the most advanced LMMs still struggle to capture aspects of compositional visual reasoning, such as attributes and relationships between objects. One solution is to utilize scene graphs (SGs)---a formalization of objects and their relations and attributes that has been extensively used as a bridge between the visual and textual domains. Yet, scene graph data requires scene graph annotations, which are expensive to collect and thus not easily scalable. Moreover, finetuning an LMM based on SG data can lead to catastrophic forgetting of the pretraining objective. To overcome this, inspired by chain-of-thought methods, we propose {\model} ({\smodel}), a novel zero-shot Chain-of-Thought prompting method that utilizes SG representations in order to extract compositional knowledge from an LMM. Specifically, we first generate an SG using the LMM, and then use that SG in the prompt to produce a response. Through extensive experiments, we find that the proposed {\smodel} approach not only improves LMM performance on several vision and language (VL) compositional benchmarks but also improves the performance of several popular LMMs on general multimodal benchmarks, without the need for fine-tuning or annotated ground-truth SGs. Code: \url{https://github.com/chancharikmitra/CCoT}.


\end{abstract}

%% file: sec/1_intro.tex
\input{figs/fig1_high_level}
\section{Introduction}
\label{sec:intro}

In recent years, \textit{Large Multimodal Models} (LMMs) such as LLaVA~\cite{liu2023llava}, GPT-4V~\cite{OpenAI2023GPT4TR}, and InstructBLIP~\cite{instructblip} have demonstrated impressive results in the field of vision and language (VL), especially in multimodal reasoning and visual question-answering (VQA)~\cite{Li2023SEEDBenchBM,Liu2023MMBenchIY,Lu2022LearnTE, antol2015vqa, Marino2019OKVQAAV}. However, recent empirical studies~\cite{Ma2022CREPE,herzig2023incorporating,doveh2022teaching} show that the best-performing VL models tend to view images as a ``bag of objects". Consider the following example in~\figgref{fig:teaser}. Suppose a VL model is asked to describe the provided image. The provided image contains many objects: a laptop, a mouse, some books, and a table. It is a challenging question to describe exactly how these objects are situated in relation to one another as well as their notable characteristics. Thus, we are motivated to utilize the SG, which captures the objects' important relationships and attributes. For example, the LMM uses the generated SG to produce the description: ``On a table, there is a stack of books on a laptop.''




Comprehending the structure of visual scenes is a core issue in machine perception. Visual scenes consist not only of objects but also include relevant characteristics and relationships that are significant to understanding the scenes' compositionality better. In this paper, we consider how to best improve the compositionality of LMMs. Recently, scene graph (SG) annotations---structured graph representations of visual scenes--have been introduced as powerful VL representations, and have been extensively explored in many previous works~\cite{sg_generation_msg_pass,herzig2018mapping,johnson2018image,Yang2022PanopticSG}. However, SG data is less readily available than textual descriptions as obtaining SGs is costly and thus not scalable.\footnote{For example, Visual Genome~\cite{krishna2017visual} contains only $\sim100K$ image-SG pairs, which is smaller than the existing LMMs pretraining datasets.} Moreover, training on SG data can lead to forgetting on the pretrained objectives as shown in~\cite{herzig2023incorporating}. Therefore, in this paper, we propose leveraging scene graph representations for LMMs \textit{without annotated scene graph data} and \textit{without finetuning}.

Recently, Large Language Models (LLMs) showed promising results by incorporating Chain-of-Thought (CoT) prompting methods~\cite{Wei2022ChainOT, Kojima2022LargeLMzshot}. CoT methods use an LLM to perform a task with intermediate reasoning steps, either zero-shot---with no explicit examples---or few-shot---with explicit examples. Inspired by this, we design a zero-shot, CoT method that utilizes scene graph representations for multimodal and compositional visual reasoning tasks. Our approach allows us to extract more \textit{compositional} knowledge out of an LMM compared to without prompting. Next, we ask ourselves how should we design a CoT prompt method that utilizes the scene graphs without relying on ground truth SG annotations or model finetuning.

Our proposed designed approach---{\model} ({\smodel})---can be broken into two steps. The first step is to generate a scene graph in order to circumvent the need for ground truth SG data by using the input image and task prompt (e.g., visual question).  The second step is to prompt the LMM with the image, task prompt, and the generated scene graph to produce a response. Incorporating the scene graph in the prompt eliminates the need for fine-tuning and prevents forgetting. Another benefit of our method is that generated SGs can describe any visual scene, therefore making {\smodel} generally applicable to a wider range of VL tasks. Finally, the fact that the generated scene graphs are compact linguistic representations of images makes {\smodel} a token-efficient prompting method. This is significant given the limited textual context lengths that LMMs often face due to processing both image and text inputs.

To summarize, our main contributions are as follows:
(i) We introduce {\smodel}, a zero-shot Chain-of-Thought approach that utilizes scene graph representations in order to extract \textit{compositional} knowledge out of an LMM;
(ii) Our proposed {\smodel} method was designed without the need for task-specific fine-tuning or annotated SG data, as well as being applicable and easy to use on various different LMM architectures;
(iii) Our method shows improved performance for LLaVA-1.5, Instruct-BLIP, SPHINX, and GPT-4V not only on VL compositional benchmarks like Winoground and WHOOPS! but also on general multimodal benchmarks like SEEDBench, MMBench, and LLaVA-Bench-in-the-Wild highlighting the effectiveness of our approach.


%% file: figs/fig1_high_level.tex
\vspace{-0.3cm}
\begin{figure}[t]
    \centering
    \includegraphics[width=1.0\linewidth]{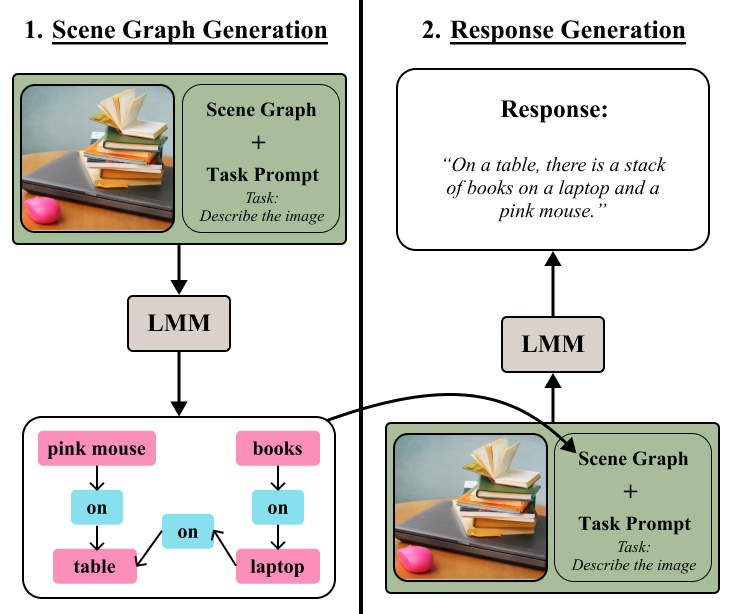}
    \caption{
    \textbf{A high-level overview of our {\model} ({\smodel}) approach.} Our {\smodel} method consists of a two-step prompting process: (1) First, the LMM is prompted to generate a scene graph relevant to the image and task prompt, such as the task in the figure ``Describe the image". (2) Following this, the LMM is prompted with the generated scene graph, the image, and the task prompt as context for responding in a way that incorporates the compositional information in the scene graph to provide a correct description of the complex scene.  
    }
    \label{fig:teaser}
    \vspace{-0.4cm}
\end{figure}


%% file: sec/2_related_works.tex
\section{Related Work}
\label{sec: rel_work}

\input{figs/fig_2_detailed}

\minisection{Large Multimodal Models (LMMs)}  The development of LMMs is largely the result of pairing LLMs' powerful reasoning capabilities~\cite{Raffel2019ExploringTL, Chowdhery2022PaLMSL,Tay2022UL2UL} with existing VL models. A good example of such models is contrastive vision and language models~\cite{radford2021clip,blip,cyclip}, which have been a significant step forward in connecting vision and language representations. However, these methods are limited in their direct application to downstream tasks that require a generative component or more explicit reasoning over both modalities, e.g., visual question-answering \cite{antol2015vqa, Hudson2019GQAAN, Saikh2022ScienceQAAN, Marino2019OKVQAAV, Jia2021ScalingUV, He2019MomentumCF}. The solution came in the form of applying the reasoning and generative capabilities of LLMs to both textual \textit{and} visual information---resulting in the development of LMMs.

LMMs directly reason over embedded visual features \cite{liu2023llava, liu2023llava15, Alayrac2022FlamingoAV, li2023blip2,instructblip, zhu2023minigpt4, Ye2023mPLUGOwlME, Ye2023mPLUGOwl2RM, Bai2023QwenVLAF, Gong2023MultiModalGPTAV, Driess2023PaLMEAE}. Particularly crucial for the success of these methods is visual instruction finetuning of the model \cite{liu2023llava, Zhao2023SVITSU}. Inspired by text-only instruction tuning of LLMs \cite{Wei2021FinetunedLM}, visual instruction tuning has been shown effective for complex visual tasks by passing detailed text descriptions and object location information to top-of-the-line LLMs (e.g. GPT-4 \cite{OpenAI2023GPT4TR}). However, this approach requires high-quality training data, which is not always available or scalable. In this paper, we present an approach that eliminates the need for training data.

Similar to LMMs, another class of multimodal methods use code generation as a proxy for visual reasoning (e.g., ViperGPT \cite{Suris2023ViperGPTVI}, VisProg~\cite{Gupta2022VisualPC}, and CodeVQA~\cite{Subramanian2023ModularVQ}), which we refer to in this paper as \textit{Visual Programmatic Models (VPMs)}~\cite{Schick2023ToolformerLM, Lu2023ChameleonPC, Shen2023HuggingGPTSA, Qin2023ToolLLMFL, Wu2023VisualCT}. Inspired by Neural Modular Network architectures~\cite{Andreas2015NeuralMN, Andreas2015DeepCQ, Johnson2017InferringAE} that leverage and scale the compositional nature of visual reasoning, VPMs build on the recent advent of highly capable out-of-the-box LLMs without the need for additional programming. Notably, these methods do not directly reason over the visual information and are limited by the exact APIs or models they are provided access to via their limited context. Unlike these methods, here we explored the potential of LMMs, which utilize scene graphs as a bridge between the visual and language domains for compositional visual reasoning.

\minisection{Multimodal Prompting Methods} Considering the growing popularity of LLMs and LMMs, prompting methods have been critical to harnessing their power as they enable precise control over model outputs and provide context within which models can be used. More importantly prompting methods occur at inference time. They include zero-shot methods~\cite{Kojima2022LargeLM, Wang2023PlanandSolvePI, Wan2023BetterZR}, few-shot methods~\cite{Brown2020LanguageMA,Min2022RethinkingTR,Dong2022ASO,Ma2023FairnessguidedFP}, expert prompting~\cite{Xu2023ExpertPromptingIL}, and Chain-of-Thought (CoT)~\cite{Wei2022ChainOT, Zhang2022AutomaticCO}, with extensions like self-consistency~\cite{Wang2022SelfConsistencyIC}, Tree-of-Thought (ToT)~\cite{Yao2023TreeOT}, and Graph-of-Thought (GoT)~\cite{Besta2023GraphOT, Yao2023BeyondCE, Lei2023BoostingLR} for more complex structures. 


To the best of our knowledge, three methods---VidIL~\cite{Wang2022LanguageMW}, DDCoT~\cite{Zheng2023DDCoTDC}, and Multimodal-CoT approaches~\cite{Zhang2023MultimodalCR, Wang2023TSciQTM}---represent the current state-of-the-art in multimodal prompting. VidIL, an architecture specifically designed for video has a language model, which reasons over captions of video frames. Similary, DDCoT designs its own CoT prompting method over image captions rather than explicit visual features. Finally, while Multimodal-CoT leverages an LMM that reasons directly over visual and text input features, its Chain-of-Thought prompting method requires finetuning on ground truth natural language reasoning, which is both annotation and computation costly.

A key difference between {\smodel} and these methods is that we utilize generated SG instead of captions (generated or collected ground-truth) as a reasoning step in our CoT design. This improves the compositionality of LMMs, which explicitly reason over visual features as well. Additionally, we demonstrate that our method enhances multimodal reasoning more broadly as well. Last, as {\smodel} is a zero-shot method used at inference time, it is broadly applicable to a wide range of LMM-based architectures.


\minisection{Compositionality} Compositionality, or the understanding of concepts as being composed of their respective subparts and relationships, is a valuable paradigm for visual concepts via reasoning over the objects, relationships, and attributes in an image. Compositionality has been applied in a variety of domains including: vision and language\cite{herzig2023incorporating,doveh2022teaching,Chen2020UNITERUI,li2020oscar,tan2019lxmert,ERNIE2021,alfassy2022feta}, visual question answering \cite{krishna2017visual, Hudson2019GQAAN, Marino2019OKVQAAV}, video understanding\cite{herzig2019stag,2020ActionGraphs,herzig2022orvit,Wang_videogcnECCV2018,avraham2022svit,materzynska2019something}, relational reasoning~\cite{baradel2018object,battaglia2018relational,Jerbi2020LearningOD}, and scene graphs\cite{johnson2015image,sg_generation_msg_pass,herzig2018mapping,herzig2019canonical,raboh2020dsg}.  Recent empirical studies~\cite{yuksekgonul2023when, vlc, winoground, herzig2023incorporating}, have shown have that even the strongest LMMs struggle to perform compositional visual understanding, including identifying object attributes and inter-object relations. Specifically, it has been shown that VL models~\cite{Ma2022CREPE} tend to learn a ``bag of objects'' representation, leading them to be less compositional. In this work, we show that a more structured CoT approach leads to improved compositional reasoning in LMMs, evidenced by improved performance on compositional benchmarks.

%% file: figs/fig_2_detailed.tex
\begin{figure*}[t]
  \centering
     \includegraphics[width=1.0\linewidth]{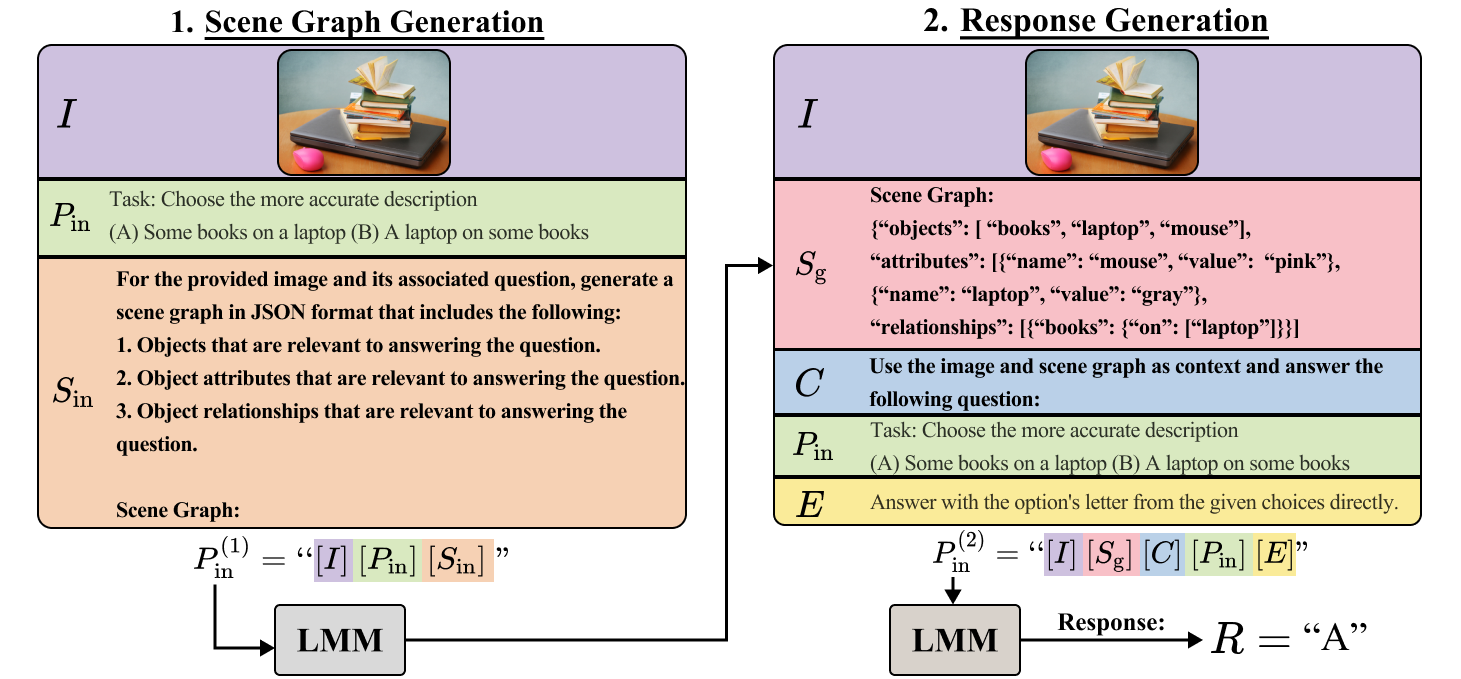}
    \caption{\textbf{Full prompt example of {\smodel}.} The first step in our prompting method is to generate a scene graph given both the image \textit{and} textual task as context. Following this, the answer is extracted by prompting the LMM with the image, scene graph, question, and answer extraction prompt. Prompt sections unique to our method are \textbf{bolded}.}
    \label{fig:full_pipeline}
\end{figure*}

%% file: sec/3_methods.tex
\section{Compositional Chain-of-Thought}
\label{sec:CCoT}

To address the challenge of LMMs viewing images as a ``bag of objects,'' as shown in previous works, our method introduces a novel approach to enhance compositional visual understanding. We begin by describing the standard LMM architecture (\Secref{sec:CCoT:preliminaries}). We then introduce our two-step chain-of-thought approach: first is scene graph generation (\Secref{sec:CCoT:sgg}) and second is response generation (\Secref{sec:CCoT:rg}). Our method is illustrated in~\figgref{fig:full_pipeline}.


\subsection{Preliminaries}
\label{sec:CCoT:preliminaries}

LMMs are multimodal models that directly reason over both vision and language modalities. They are typically given inputs of one image $I$ and an associated task prompt in text form $P_{\mathrm{in}}$ (e.g., questions, caption generation, etc.). 
Each modality is then encoded into a shared embedding space that a language model $f_{\theta}(\cdot)$ (parameterized by $\theta$) can reason over. More concretely, the image is encoded using a trainable vision encoder $v_{\phi}(\cdot)$ (parameterized by $\phi$), while the task prompt is tokenized and then encoded using a fixed language embedding $l$. Given an input image $I$ and input task prompt $P_\mathrm{in}$, the language model (typically an LLM) then outputs a text response $R$.
\nolinebreak
\begin{equation}
    R = f_\theta(v_\phi(I), l(P_\mathrm{in}))
\end{equation}

The exact LMM sub-modules of the LLM, vision encoding architecture, and pretraining method for parameters ${\theta, \phi}$ differ between models but the overarching method described above remains the same. 

We propose {\smodel}, a zero-shot chain-of-thought prompting method that leverages scene graph generation to improve an LMM's compositional visual understanding and multimodal reasoning.  Notably, this method does not require any finetuning as it is purely prompting-based. Furthermore, no annotated SGs are required as the method is zero-shot. Ultimately, our method is centered around a scene-graph generation prompt $S_\mathrm{in}$ that can be integrated into $P_\mathrm{in}$ such that the LMM can output a scene graph $S_\mathrm{g}$ as an intermediate multimodal reasoning step to output better responses to the task prompts, such as questions, classification, or caption generation. 


\subsection{Step 1: Scene Graph Generation}
\label{sec:CCoT:sgg}

Our first step is to generate a scene graph $S_\mathrm{g}$, obviating the need for ground truth annotated SG data. The scene graph generation prompt $S_\mathrm{in}$ instructs the LMM to systematically construct a scene graph with three key properties: the \textit{objects}, their \textit{attributes}, and the \textit{relationships} between them. To address the ``bag-of-objects'' problem, we would like to have a global view of not just the objects, which are the primary units for visual reasoning, but also their properties and how they interact with one another.

In the scene graph generation prompt $S_\mathrm{in}$, we further condition its format to be in JSON. This standardization in JSON format is intended to facilitate easier interpretation by the LMM. By systematically organizing visual information through the inclusion of objects, relationships, and attributes in the scene graphs, we enable more structured and comprehensive reasoning. The full prompt, showcasing this structured approach, is illustrated in Figure~\ref{fig:full_pipeline}. The scene graph generation method represents a core novel contribution of our work, aiming to overcome the limitations of existing multimodal reasoning models and enhance the compositional understanding of LMMs.


We include both the image $I$ and task prompt $P_\mathrm{in}$ along with $S_\mathrm{in}$ to condition the generated scene graph to be relevant to the given task prompt. This is because SGs are inherently very long-tailed: a generated scene graph that is conditioned only on the image, might incorporate information unrelated to the given task prompt.

The entire first prompt to the LMM, which we denote as $P_\mathrm{in}^{(1)}$ is constructed by combining the input image $I$, task prompt $P_\mathrm{in}$, and most notably the scene-graph generation prompt $S_\mathrm{in}$ (showed in red under Scene-Graph Generation in Figure~\ref{fig:full_pipeline}). The full prompt is as follows:
\nolinebreak
\begin{equation}
     P_\mathrm{in}^{(1)} = \text{``[$I$] [$P_\mathrm{in}$] [$S_\mathrm{in}$]"}
\end{equation} 
where $[\cdot]$ indicates slots for inserting the individual elements of the prompt. The LMM thus generates a SG as follows:
\begin{equation}
    S_\mathrm{g} = f(v_\phi(I), l(P_\mathrm{in}^{(1)}))
\end{equation}


\subsection{Step 2: Response Generation}
\label{sec:CCoT:rg}

To bypass the need for finetuning and thus eliminate forgetting, we utilize the generated scene graph $S_\mathrm{g}$ as an intermediate chain-of-thought reasoning step.
The LMM is thus prompted with the original task prompt, image, and corresponding generated scene graph so that all three can be jointly used as context to respond to this new task prompt. The overall input prompt for response generation is thus given as follows:
\nolinebreak
\begin{equation}
    P_\mathrm{in}^{(2)} = \text{``[$I$] [$S_\mathrm{g}$] [$C$] [$P_\mathrm{in}$] [$E$]"}
\end{equation}

In addition to the input image $I$, original task prompt $P_\mathrm{in}$, and generated scene graph $S_\mathrm{g}$, we insert a context sentence $C$ and an answer extraction sentence $E$. $C$ briefly instructs the LMM to use the provided context. Concretely, this is given by ``Use the image and scene graph as context and answer the following question:''. Finally, while the flexibility of LLM text generation is a great modeling choice for high-level multimodal reasoning, this flexibility also makes response generation in a specific format non-trivial. Many multimodal benchmarks are in a multiple-choice format, for example. Since we evaluate our method on these types of benchmarks, a short additional sub-prompt $E$ (usually a conditioning sentence) is required to return the answer as a letter. For example, our answer extraction sub-prompt ``Answer with the option's letter from the given choices directly" is taken from LLaVA-1.5~\cite{liu2023llava15} as it has been shown to be reliable on large multiple-choice benchmarks. However, this method can be easily generalized to other answer formats like short answers or detailed descriptions by modifying or completely removing $E$. Thus, the LMM generates a final response  $R$ to the original image, task prompt pair $(I, P_\mathrm{in})$ as follows:
\begin{equation}
    R = f(v_\phi(I), l(P_\mathrm{in}^{(2)}))
\end{equation}


%% file: sec/4_evaluation.tex
\input{tables/tbl_1}

\section{Experiments and Results}

We apply our {\smodel} approach to four popular LMMs: InstructBLIP-13B~\cite{instructblip}, LLaVA-1.5-13B~\cite{liu2023llava15}, Sphinx~\cite{Lin2023SPHINXTJ}, and GPT-4V~\cite{OpenAI2023GPT4TR}. We also evaluated our approach to several baselines across different benchmarks, focusing on multimodal reasoning and VL compositional tasks. Additional results can be found in our Supplementary~\Secref{supp:expr}.

\subsection{Implementation Details}
\label{sec:eval:impl}

We implemented {\smodel} using PyTorch~\cite{paszke2019pytorch}. In order to obtain pre-trained models which we evaluated our method, we used each model's respective official implementation. While the compute and memory requirements differ between models, our prompting method needs only the infrastructure necessary for running inference on these models. Refer to Supplementary in~\Secref{supp:impl} for more information.


\input{tables/tbl_2}
\subsection{Datasets}

The goal of our work is to demonstrate that our method improves LMMs' compositional visual understanding, while also enhancing a broad range of vision-and-language tasks. 
In what follows next, we describe our evaluation datasets.


\minisection{VL Compositional Benchmarks} To evaluate the compositional visual understanding of our method, we consider the Winoground~\cite{winoground} and WHOOPS!~\cite{BittonGuetta2023BreakingCSWHOOPS} benchmarks: \textbf{(1) Winoground} is a hand-curated dataset designed to test VL models' compositional visual understanding. Each sample contains two images and a corresponding pair of image captions. Both captions are syntactically very similar but contain one key difference in the form of a semantic swapping of objects, relations, or both. On the same dataset, Winoground performance is evaluated on three metrics: (i) a text score, where the correct caption must be identified given one image; (ii) an image score, where the correct image must be identified given one caption; (iii) a group score, where the two pairs must be matched correctly. \textbf{(2) WHOOPS!} similarly tests compositionality using images that violate typical visual commonsense. There are a broader variety of tasks, in particular: (i) Explanation Generation, (ii) Image Captioning, (iii) Cross-Modal Matching, and (iv) Compositional VQA. We evaluate our method on the Compositional VQA split of the dataset.

\minisection{Multimodal Reasoning Benchmarks} Recently, there has been an introduction of several new benchmarks that are specifically designed to evaluate the multimodal reasoning abilities of LMMs. In our work, we focus on SEEDBench~\cite{Li2023SEEDBenchBM}, MMBench~\cite{Liu2023MMBenchIY}, and LLaVA-Bench In-the-Wild~\cite{liu2023llava15}. Both SEEDBench and MMBench include different splits that test general visual perception and visual reasoning. For instance, SEEDBench contains perception tasks that evaluate an LMM's Instance Identification and Instance Attribute understanding capabilities while also containing more higher-order reasoning splits like Scene Understanding and Instance Interaction. MMBench has similar splits. We exclude video, thus evaluating our method on the image splits of SEEDBench and the entirety of MMBench. To evaluate a different type of multimodal reasoning, we further evaluate our method on LLaVA-Bench In-the-Wild, which tests the LMMs' ability to give detailed long-form answers to visual questions.


\subsection{Models} 
\label{sec:eval:models}

In our work, we apply our {\smodel} approach to four popular LMMs described as follows.


\minisection{LLaVA-1.5} The LLaVA~\cite{liu2023llava} architecture distinguishes itself as a powerful state-of-the-art (SOTA) LMM method. 
Featuring a simple linear projection that maps CLIP visual features of the input image into a shared embedding space with the LLM language tokens, LLaVA instruction tunes on a dataset of images--LLaVA-Instruct-158k---paired with conversational, detailed description, and complex reasoning response types for better visual alignment than simply image-text pairs. 
In our work, we evaluate LLaVA-1.5~\cite{liu2023llava15}, a newer version of LLaVA with improved baselines. Model improvements over the original architecture include: (1) replacing the linear projection with an MLP and (2) pretraining on more diverse datasets.


\minisection{InstructBLIP} While InstructBLIP also uses a frozen visual encoder and LLM, it calculates visual features via a Q-former transformer as in BLIP-2~\cite{li2023blip2} model that outputs learnable visual tokens. The difference, in this case, is that InstructBLIP's Q-former also attends over the task prompt, making the visual features \textit{instruction-aware}. This, in addition to a broader set of visual instruction tuning datasets that includes the LLaVA-Instruct-158k affords the method high performance on benchmarks like SEEDBench~\cite{Li2023SEEDBenchBM}.

\minisection{SPHINX} Sphinx~\cite{Lin2023SPHINXTJ} distinguishes itself from other LMMs in two key ways: Sphinx (1) unfreezes its LLM weights during instruction finetuning and (2) has a broader area of multimodal question-answering tasks including "region-level understanding, caption grounding, document layout detection, and human pose estimation''~\cite{Lin2023SPHINXTJ}. 

\minisection{GPT-4V} Unlike the other three models, GPT-4V's architecture and pretraining details are not made public. However, using the SOTA GPT-4 as the LLM backbone will be essential in evaluating how our method works on an LMM with superior language reasoning skills.

\input{tables/tbl_5}


\subsection{Baselines}
In our experiments, we compare our {\smodel} prompting methodology to two other prompting baselines as shown in Table~\ref{tbl: main}. First, to evaluate the added benefit of our method to pretrained LMMs, our first baseline is to apply the model to the benchmark without any prompt engineering. Second, we consider a baseline of a \textit{language} zero-shot (ZS) CoT prompting method \cite{Kojima2022LargeLMzshot} to determine the benefit of {\smodel} compared to a SOTA CoT prompting method. The method works in a two-step fashion. (i) Given the input question and text, the reasoning trigger ``Let's think step-by-step.'' is appended to the end of the prompt, coming subsequently after the question. This generates language reasoning for an answer to the question. (ii) Because the answer is implicit in this outputted reasoning, the second step involves passing the image, question, output reasoning from step 1, and an answer extraction phrase to return a response in the desired format. We find that compared to the answer extraction phrase suggested in the original paper, the one suggested by LLaVA~\cite{liu2023llava15} yields higher accuracy on most benchmarks and so proceed with this slight change compared to the original implementation of ZS-CoT. We also compare our work to the recent SOTA multimodal CoT prompting methods MMCoT~\cite{Zhang2023MultimodalCR}, DDCoT~\cite{Zheng2023DDCoTDC}, and VidIL~\cite{Wang2022LanguageMW} on the SEEDBench-Image dataset as shown in Table~\ref{tbl: baselines}

%% file: tables/tbl_1.tex
\renewcommand{\arraystretch}{1.1}
\newcolumntype{?}{!{\vrule width 2 pt}}
\newcolumntype{P}[1]{>{\centering\arraybackslash}p{#1}}
\newcolumntype{G}[1]{>{\columncolor{lightgray}\centering\arraybackslash}p{#1}}
\begin{table*}[t]
\begin{center}
\begin{tabular}{m{0.22\textwidth}G{0.08\textwidth}G{0.08\textwidth}G{0.09\textwidth}P{0.08\textwidth}P{0.08\textwidth}P{0.08\textwidth}P{0.08\textwidth}}
        \multicolumn{1}{c}{}& \multicolumn{3}{c}{\textbf{Multimodal Benchmarks}} & \multicolumn{4}{c}{\textbf{VL Compositional Benchmarks}} \\
        \toprule
        Model & SEED-I & MMBench  & LLaVA-W& Wino-Text & Wino-Image &Wino-Group & WHOOPS! VQA BEM  \\ \hline
        CLIP & - & -&-& 30.7 & 10.5 & 8.0 & - \\
        BLIP & - & -&-& 39.0 & 19.2 &  15.0 & 39.0 \\
        BLIP2 & 46.4 & - &-& 42.0 & 23.8 & 19.0 & 55.0 \\
        SGVL$^\dag$ & - &  -&-& {42.8}\dag & {28.5}\dag & {23.3}\dag & -  \\
        mPlug-OWL2 & 57.8 &  64.5&-& - & - & - & -  \\
        QwenVL-Chat & 58.2 &  61.2&-& - & - & - & -  \\ \hline
        InstructBLIP-13B & 48.2 &  36.0 &47.2& 12.8 & 13.3& 4.5  & 48.3\\ 
        InstructBLIP-13B-ZS-CoT & 37.6 & 25.3 &45.4& 15.8 & 14.8 & 6.0 & 43.36 \\
        \textbf{InstructBLIP-13B-{\smodel}} & 56.9\gcol{+8.7} &40.3\gcol{+4.3}&47.9\gcol{+0.7}&26.0\gcol{+13.2} & 27.0\gcol{+13.7} & 11.5\gcol{+7.0} & 62.9\gcol{+14.6}\\ \hline
        LLaVA-1.5-13B & 68.2 &  67.0 &73.5& 33.5 & 35.0 & 17.3 & 47.3  \\
        LLaVA-1.5-13B-ZS-CoT & 66.7 &66.0 &68.5&36.8& 35.0&19.8&46.6\\
        \textbf{LLaVA-1.5-13B-{\smodel}} & 69.7\gcol{+1.5} & 70.7\gcol{+3.7}&74.9\gcol{+1.4}&39.8\gcol{+6.3} & 37.3\gcol{+2.3}&22.3\gcol{+5.0} & 61.2\gcol{+13.9}  \\ \hline
        Sphinx & 71.6 & 65.9 &70.0& 29.0 & 29.0&  16.3& 50.0\\ 
        Sphinx-ZS-CoT & 70.3 & 65.5 &69.8& 36.0 & 38.5 & 21.5 & 60.4 \\
        \textbf{Sphinx-{\smodel}} & 74.2\gcol{+2.6}  &68.3\gcol{+2.4}&71.0\gcol{+1.0}&36.5\gcol{+7.5} & 36.3\gcol{+7.3} & 22.5\gcol{+6.2} & 61.9\gcol{+11.9}
        \\ \hline
        GPT4V & 69.1 & 75.5 &88.2& 60.3 & 45.3&  33.5&64.8\\ 
        GPT4V-ZS-CoT & 72.5 & 74.8 &88.8& 63.3 & 52.5 & 41.0 & 65.5 \\
        \textbf{GPT4V-{\smodel}} & 74.0\gcol{+4.9}  &76.3\gcol{+0.8} &91.2\gcol{+2.0}& 64.0\gcol{+3.7} & 54.5\gcol{+9.2} & 43.3\gcol{+9.8}&67.8\gcol{  +3.0}
        \\\bottomrule
    \end{tabular}
    \end{center}
\caption{\textbf{Main results table on SeedBench, MMBench, Winoground, and WHOOPS! Benchmarks.} Abbreviations: SEEDBench-Image [SEED-I]; Winoground Text Score: Wino-Text, Image Score: Wino-Image, Group Score: Wino-Group. Unlike our zero-shot approach, models with \dag\ are supervised and finetuned on annotated scene graphs. For more results, please refer to~\Secref{supp:expr:more_results} in Supp.}
\label{tbl: main}
\end{table*}

%% file: tables/tbl_2.tex
\renewcommand{\arraystretch}{1}
\newcolumntype{?}{!{\vrule width 2 pt}}
\newcolumntype{P}[1]{>{\centering\arraybackslash}p{#1}}

\begin{table*}[h!]
\begin{center}
\begin{tabular}{m{0.21\textwidth}P{0.045\textwidth}P{0.045\textwidth}P{0.045\textwidth}P{0.045\textwidth}P{0.045\textwidth}P{0.045\textwidth}P{0.045\textwidth}P{0.045\textwidth}P{0.045\textwidth}G{0.1\textwidth}}
        \toprule
        Model & IC & SU & IId & IA & IL & SR & VR & TU & IIn & Overall \\
        \hline
        MMCoT\dag & 22.1 & 29.5 & 30.2 & 32.8 & 33.6 & 30.3 & 34.1 & 45.9 & 34.0 & 34.4 \\
        LLaVA-1.5-13B-DDCoT & 47.3 & 63.0 & 59.8 & 64.1 & 44.6 & 41.4 & 67.1  & 57.7& 51.6 & 58.0\\
        LLaVA-1.5-13B-VidIL & 62.3 & 74.9 & 72.5 & 69.9 & 62.5 & 53.9 & 78.0  & 49.4& 71.1 & 68.9\\
        \textbf{LLaVA-1.5-13B-{\smodel}} & 59.3 & 76 & 74.4 & 71.8 & 64.3 & 54.5 & 79.2  & 58.8& 74.2 & 69.7\\ \bottomrule
    \end{tabular}
    \end{center}
    
\vspace{-3mm}
\caption{\textbf{Comparison to Multimodal CoT Methods.} TBD Instances Counting [IC], Scene Understanding [SU], Instance Identity [IId], Instance Attributes [IA], Instance Location[IL], Spatial Relation [SR], Visual Reasoning [VR], Text Understanding [TU], Instance Interaction[IIn]. Note that \dag indicates that MMCoT is a finetuning method that was pretrained on ScienceQA.}
\label{tbl: baselines}
\end{table*}

%% file: tables/tbl_5.tex
\renewcommand{\arraystretch}{1}
\newcolumntype{?}{!{\vrule width 2 pt}}
\newcolumntype{P}[1]{>{\centering\arraybackslash}p{#1}}

\begin{table*}[htbp]
\begin{center}
\begin{tabular}{m{0.25\textwidth}P{0.05\textwidth}P{0.05\textwidth}P{0.05\textwidth}P{0.05\textwidth}P{0.05\textwidth}P{0.05\textwidth}P{0.05\textwidth} G{0.07\textwidth}}
        \toprule
        Model  & SU & IId & IA & IL & SR & VR &  IIn & W. Avg. \\ \midrule
        \textbf{LLaVA-1.5-13B-{\smodel}} & 76.0 & 74.4 & 71.8 & 64.3 &  54.5& 79.2 & 74.2 & 72.1\\ 
        LLaVA-1.5-13B  & 74.9 & 71.3 &68.9 & 63.5 & 51.5 & 77.0& 73.2&69.9 \\
        \midrule
        \quad w/ Object Locations & 75.4 & 72.7 & 69.4 & 63.6 & 54.5 & 78.9 & 73.2 & 70.5\\ 
        \quad w/out JSON Format & 74.8 & 73.1 & 70.7 & 63.0 & 52.0 & 78.6 &  73.2 & 68.1 \\
        LLaVA-1.5-13B-Caption-CoT  & 75.7 & 73.1 &69.1 & 63.1 & 55.3 & 78.6 & 73.7 & 70.7 \\ 
        LLaVA-1.5-7B  & 50.6 & 42.2 & 43.0 & 38.1 & 33.8 & 58.0 & 50.5 & 66.3\\ 
        LLaVA-1.5-7B-CCoT & 68.7 & 57.9 & 63.7 & 47.9 & 42.8 & 67.1 & 66.0 &66.1\\ 
        \midrule
        128 Token Length & 76.2 & 73.4 & 71.4 & 63.7 & 55.4 & 80.1  & 75.3 &  71.9\\
        512 Token Length & 75.5 & 73.6 & 71.6 & 63.2 & 54.8 & 79.15 & 74.2 & 71.6\\
        1024 Token Length & 75.9 & 73.5 & 71.7 & 63.2 & 54.0 & 79.5  & 76.3 &  71.5\\

        \bottomrule
        
    \end{tabular}
        
\end{center}
\caption{\textbf{Ablations on SEEDBench-Image.} This table describes key split-level ablation results of our method on all image splits of SEEDBench~\cite{Li2023SEEDBenchBM}: Instances Counting [IC], Scene Understanding [SU],Instance Identity [IIn], Instance Attributes [IA], Instance Location[IL], Spatial Relation [SR], Visual Reasoning [VR], Text Understanding [TU], Instance Interaction[IIn]. W. Avg. denotes the weighted average.}
\label{tbl: ablations}
\end{table*}

%% file: sec/5_results.tex
\subsection{Results}
\label{sec: results}


Results are shown in~\tabref{tbl: main}. An advantage of our method is that it can be applied across a variety of different pretraining methods and visual architectures. We demonstrate that applying {\smodel} outperforms the base models across several benchmarks, highlighting the effectiveness of our approach. In Figure~\ref{fig:examples}, we show concrete examples where our method improves upon baselines as well as cases where it still fails. For more results, refer to~\Secref{supp:expr:more_results} in Supplementary.

\minisection{Compositional visual understanding} 
For all four LMMs tested, we find substantial increases utilizing {\smodel} compared to baselines when evaluated on Winoground and WHOOPS! In fact, without any instruction tuning, GPT-4V-{{\smodel}} achieves a significant improvement over the previous Winoground SOTA---SGVL, which has been finetuned on ground truth SG annotations~\cite{herzig2023incorporating}. Interestingly, ZS-CoT method actually \textit{degrades} performances across several splits of the compositional benchmarks. This may be due to the lack of consideration for visual information in the prompt as it was designed for LLMs. Thus, these results demonstrate the effectiveness of {\smodel} for improving compositional visual reasoning of LMM without the need for finetuning or ground-truth annotated SG data.


\minisection{Multimodal Benchamarks}
We also see that {\smodel} improves over the baselines on SEEDBench image splits, MMBench, and LLaVA-Bench In-the-Wild. Even with many LMMs having a variety of different LLM backbones and pretraining methods, the difference between consecutive state-of-the-art models on SEEDBench is usually 1\% or less. All of {\smodel}'s improvements are 1\% or better. Therefore, these results are a robust indication that our method is advantageous for elevating both LMMs' compositional visual understanding and their general multimodal reasoning. Once again, ZS-CoT prompting is actually often detrimental to LMMs on many splits of these benchmarks. 



\input{figs/fig_3_examples}
\subsection{Ablations}

We perform a comprehensive ablation study on SEEDBench with our LLaVA-1.5-{\smodel} model (see \tabref{tbl: ablations}). We note that we did not report the Instance Counting and Text Understanding (OCR) splits as they do not constitute visual reasoning. For more ablations,
refer to \Secref{supp:more_ablt} in Supp.  

\minisection{Requiring bounding boxes} In our qualitative exploration of generated SGs, we found that some SGs included bounding-box coordinates for objects. Thus, we experimented with a prompt that instructed the LMM to include bounding-box coordinates (shown in the table as``w$\backslash$ Object Locations'') for all objects in the generated SG. We find a 1.6\% decrease in weighted average accuracy on SEEDBench-Image suggesting that requiring exact object locations is not beneficial to multimodal reasoning tasks.

\minisection{JSON structure enhances SG utilization}
While SGs are structured visual representations, they may come in many different textual formats. As such, we ablate the JSON format requirement (refer to as \textit{w/out JSON Format}) of our SG generation prompt to evaluate whether enforcing a specific SG format affects the LMMs usage of the content. Our results indicate that enforcing a common, systematic format like JSON is indeed beneficial (-2.0\% without JSON) to the LMMs ability to most effectively utilize the SG.

\minisection{Replacing SGs with captions}
SGs are a \textit{highly-structured} representation of visual information that distinguishes them from simply natural language descriptions of images. Therefore, we ablate the importance of SG structure by generating captions instead of SGs (refer to as \textit{LLaVA-1.5-Caption-CoT}). We find in Table~\ref{tbl: ablations} that generating captions with the \textit{same informational context as our SG method}, but, degrades performance (-1.4\% compared to ours), suggesting the importance of SG structure to multimodal tasks. 

\minisection{LMM size} We also evaluate the impact of the LMM size. We find that LLaVA-1.5-7B-\smodel\ shows no noticeable difference (+.1 \%) in accuracy compared to LLaVA-1.5-7B. The more substantial gains of LLaVA-1.5-13B-\smodel\ and GPT-4-\smodel\ indicate that our method is most effective for larger model sizes. This facet is crucial as our zero-shot method becomes a comparatively less compute-expensive process than the finetuning for these larger LMMs.

\minisection{Effect of SG size}
We consider how the size of the SG affects the generated response, by comparing the accuracy when using SGs of different token lengths. Concretely, we evaluate when using SGs of length 1024 (-.6\%), 512 (-0.5\%), and 128 (-0.3\%) tokens. The results demonstrate that the optimal SG size is 256 tokens. This demonstrates the effectiveness of textual SGs in encapsulating useful information in a small sequence length while also providing credence to the idea that a minimum amount of information is necessary to properly respond to the question.



%% file: figs/fig_3_examples.tex
\begin{figure*}[t]
  \centering
     \includegraphics[width=1.0\linewidth]{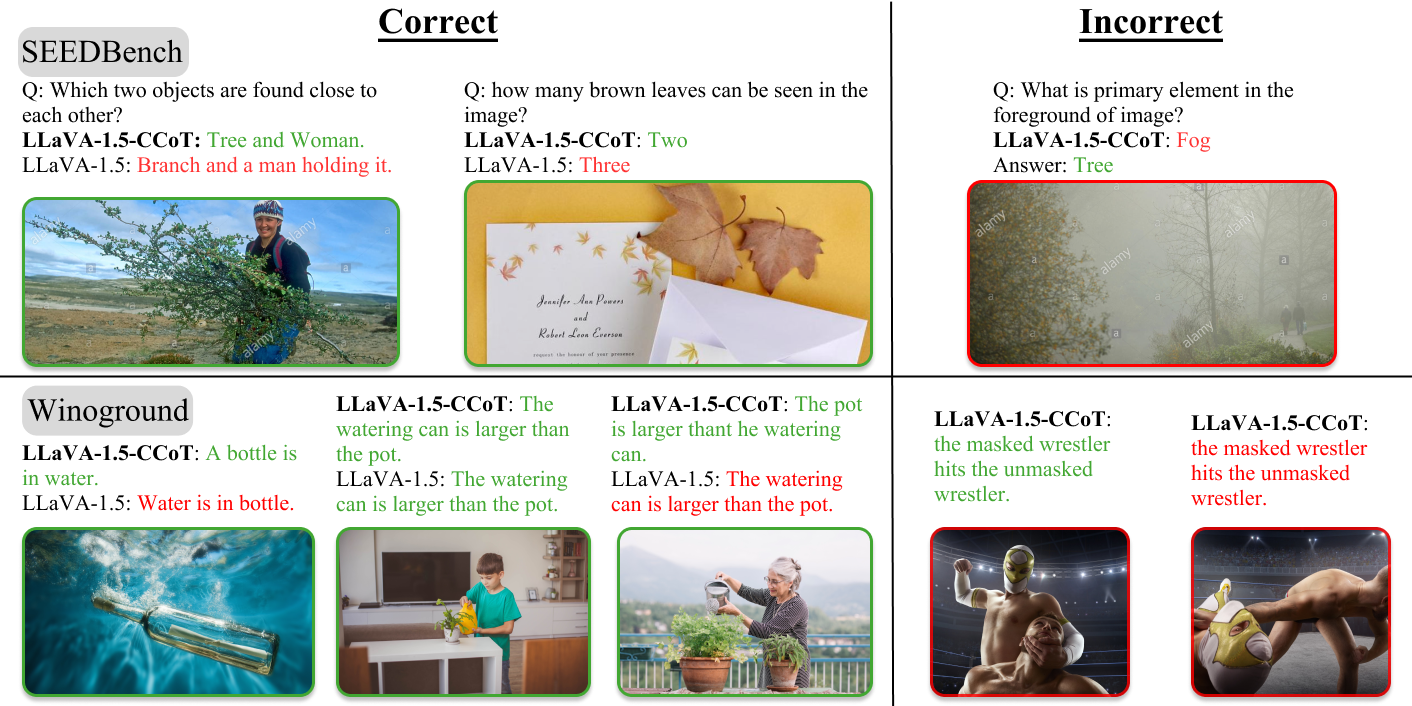}
    \caption{\textbf{Example Outputs.} Above we show examples of our method on both SEEDBench and Winoground. On the left we show successes of {\smodel} while the right shows failure cases. For more qualitative visualizations, please refer to~\Secref{supp:qual} in Supplementary.} 
    \label{fig:examples}
\end{figure*}

%% file: sec/6_conclusion.tex
\section{Conclusion}
Our research has demonstrated the significant potential of the {\smodel} approach in extracting compositional information from an LMM. This extracted knowledge leads to enhanced compositional visual and multimodal reasoning of LMMs downstream without the need for fine-tuning or reliance on ground-truth annotated SG data. Our method stands out by generating SGs in a zero-shot manner, effectively addressing the issue of annotated SG availability. Using the generated SG in a CoT reasoning prompt also addresses catastrophic forgetting by not fine-tuning. The substantial improvements observed on compositional visual reasoning benchmarks like Winoground and WHOOPS!, along with the general multimodal benchmarks SEEDBench, MMBench, and LLaVA-Bench In-the-Wild underscore the effectiveness of our approach across a diverse set of tasks. This is further corroborated by our ablations, which reveal the importance of using structured SGs over captions, leveraging the JSON format, and utilizing optimal SG length to enhance the LMMs' visual compositional and multimodal reasoning. These results collectively highlight the value of our method in broadening the capabilities of LMMs
in compositional and multimodal reasoning tasks.

%% file: sec/7_impacts.tex
\section{Limitations} 

In this work, we present a zero-shot Chain-of-Thought prompting method that utilizes scene-graph representations for multimodal and compositional visual reasoning tasks. We demonstrate improved performance on several different models and benchmarks. Nevertheless, our work has a central limitation. While extending context length is an active field of research, our method is limited by the current context lengths of the LLMs being used by the LMMs. Additionally, scene graphs are not particularly useful representations when performing multimodal tasks that emphasize language over visual reasoning, such as document understanding. Finally, we do not anticipate negative impacts of this work, but, as with any machine learning method, we recommend exercising caution.






%% file: sec/supp.tex

\clearpage
\setcounter{page}{1}
\maketitlesupplementary

Here we provide additional information about our experimental results, qualitative examples, implementation details, and datasets. Specifically, \Secref{supp:expr} provides more experiment results, \Secref{supp:impl} provides additional implementation details, and \Secref{supp:qual} provides qualitative visualizations to illustrate our approach.

\input{tables/tbl_7_supp}
\input{tables/tbl_3}
\input{tables/tbl_4}

\input{tables/tbl_6_supp}

\section{Additional Experiment Results}
\label{supp:expr}

We begin by presenting several additional ablations (\Secref{supp:more_ablt}) that further demonstrate the benefits of our {\smodel} approach. We also present additional results (\Secref{supp:expr:more_results}) on MMBench Perception Splits.

\subsection{Additional Ablations}
\label{supp:more_ablt}

In what follows, we provide additional ablations that further illustrate the benefits of {\smodel}. For all ablations, we compare the ablated experiment with the corresponding best-performing {\smodel} results on the SEEDBench-Image dataset.

\minisection{Random SG Regularization}
To assess whether only the structure of SGs can be valuable for reasoning, we ablate the specific details of the scene graph. Concretely, we pass a random SG not related to the input question to the LMM and prompt the model to only use it as a framework for reasoning, essentially regularizing the response via the random SG. We find a 5.6\% degradation in performance from \textbf{LLaVA-1.5-{\smodel}} as shown in Table~\ref{tbl: ablations_supp}. The ablation demonstrates that providing the structure of an SG without accurate content leads to an accuracy decrease compared to LLaVA-1.5-\smodel\ (-5.6\%). The capability to generate correct answers despite suboptimal reasoning steps is supported by recent literature---e.g. CoT~\cite{Wei2022ChainOT} and DDCoT~\cite{Zheng2023DDCoTDC}. This demonstrates that our method does not require ground truth SGs but also generates SGs accurately enough to make LMMs substantially more effective on compositional \& multimodal tasks.

\minisection{SG Knowledge Distillation}
Here, we ask whether the compositional knowledge extracted in high-quality scene graphs can be transferred to other models. In particular, we use \textbf{LLaVA-1.5-13B-{\smodel}} scene graphs when prompting InstructBLIP on SEEDBench-Image splits. The entire process for running \textbf{InstructBLIP-13B-{\smodel}} remains the same with just the scene graph being replaced by the one generated by \textbf{LLaVA-1.5-13B-{\smodel}}. We find that this actually leads to a slight degradation in performance of \textbf{InstructBLIP-13B-{\smodel}}, but still better than InstructBLIP-ZS-CoT. 

\minisection{Removing Image}
Finally, we ablate the effect of removing the image during the second response generation step. The first step of our method is left intact, using the image and question as context to yield a relevant scene graph. However, in the second step, we generate a response with the image tokens masked. This experiment evaluates the gap between the SG and the visual information offered by the image. We find a significant decrease in performance of 26.1\%, indicating that it is a combination of the image and scene graph that affords our method its improved performance over baselines.


\minisection{Impact of individual compositional characteristics} We individually ablate the objects, attributes, and relationships of the generated \smodel\ SGs. We find improvements in accuracy over LLaVA-1.5 when we remove the objects/attributes/relationships from LLaVA-1.5-13B-\smodel\ by 1.5/0.8/1.2 \% on SEEDBench and 3.5/3.5/2.25 \% on Winoground Text. These results indicate that each of the three elements of \smodel\ contributes to its effectiveness, but combined, they achieve the best result.

\minisection{Removing the term "Scene Graph" from prompt} It is possible that the term "scene graph'' encapsulates some latent understanding of visual compositionality that the LMM has already. To evaluate this, we ablate the term "scene graph'' from our prompt and simply replace it with the word "description''. Indeed this shows a -2.1\% decrease in accuracy on the evaluated SEEDBench-Image splits. This suggests that LMMs have some knowledge of SGs from pretraining that is helpful when generating a structured compositional representation to aid in multimodal reasoning.

\minisection{COCO and Visual Genome Data Overfitting} Most LMMs have been extensively instruction-tuned on images sourced from the COCO~\cite{Lin2014MSCOCO} and Visual Genome~\cite{krishna2017visual} datasets. In fact, LLaVA-Instruct-158k, the main dataset used to finetune LLaVA and many other LMMs consists solely of COCO images in order to make use of the bounding box and detailed description information provided by the dataset. Besides this, there are many instances of other datasets that reuse COCO and Visual Genome images. As such, our method is effective in helping these LMMs generalize to datasets like SEEDBench, MMBench, and Winground, while showing no substantial difference in performance compared to the zero-shot case on benchmarks like GQA~\cite{Hudson2019GQAAN} and VL Checklist~\cite{vlc} that heavily make use of the overfit COCO or Visual Genome images (CCoT is -.8 \% and -.6 \% compared to zero-shot on GQA and VL Checklist respectively).

\subsection{Additional Results}
\label{supp:expr:more_results}

\minisection{Detailed Split Results} We present detailed results of our method on the individual SEEDBench-Image splits as well as all of MMBench's splits which are separated into their Perception and Reasoning categories. These results shown in Tables~\ref{tbl: detailedSEED},~\ref{tbl: MMBReasoning}, and~\ref{tbl: MMBenchPerception_supp}.



\minisection{LLaVA-Bench Results} LLaVA-Bench is a challenging hand-designed dataset on a small number of images and questions for evaluating the effectiveness of LMMs as multimodal chatbots. The questions are designed in a more open-ended conversational manner, differing greatly from the other benchmarks presented in the main paper.  Instead of evaluating a simple response to a question, the benchmark tests the entire long-form text conversions, meaning there is a greater burden to account for the language response. Examples of LLaVA-Bench are shown in Figure~\ref{fig:examples}. Even so, we show slight improvements over the baselines--+0.4\% for InstructBLIP and +0.3\% for LLaVA-1.5--when ZS-CoT shows significant degradation (roughly 10\% decrease). This suggests that our method is also potentially beneficial for more open-ended visual chat applications, which is different than multimodal reasoning and VL compositional benchmarks.



\section{Additional Implementation Details}
\label{supp:impl}

To run our models on larger benchmarks, we use 8 NVIDIA RTX 6000 GPUs to split the datasets across multiple GPUs. Smaller datasets like Winoground and MM-Vet are able to run individual experiments on a single GPU. Besides the output token generation length, we use the default generation parameters (e.g. temperature and no. of beams in beam search) for each model. For any baseline performance already reported by the official codebase of the model (e.g. LLaVA-1.5 on SEEDBench or MMBench), we use the value presented in that model's corresponding paper. Please refer to the respective model's paper for their specific implementation details of the architecture. In the following sections, we describe some nuances of our method applied on different datasets.

\subsection{SEEDBench}
\minisection{Dataset}
SEED-Bench~\cite{Li2023SEEDBenchBM}is a large-scale benchmark designed to provide a comprehensive and objective evaluation of LMMs, particularly focusing on generative comprehension. This benchmark comprises 19k multiple-choice questions, all of which have been annotated by humans. These questions are structured to cover 12 evaluation dimensions, catering to both spatial and temporal understanding.

The development of SEED-Bench involved designing an advanced pipeline specifically for creating multiple-choice questions. This pipeline is tailored to target specific evaluation dimensions, thereby enabling the scalability of evaluation data across various domains. Furthermore, the benchmark incorporates a blend of automatic filtering and manual verification processes to enhance the quality of the generated questions and answers.

For the specific research paper in question, only the image splits of SEED-Bench are utilized for evaluation purposes. 

\minisection{Inference details} We use the official dataset released by the authors which is available at \url{https://github.com/AILab-CVC/SEED-Bench}. All models evaluated on SEEDBench use the exact method described in Section~\ref{sec:CCoT} of the main paper.

\subsection{MMBench}
\label{supp:impl:mmbench}

\minisection{Dataset} MMBench \cite{Liu2023MMBenchIY} is a novel multimodal benchmark created to address the limitations of existing benchmarks like VQAv2 \cite{antol2015vqa} or COCO Captions\cite{Lin2014MSCOCO}, which provide quantitative performance measurements but lack fine-grained ability assessment and robust evaluation metrics. Unlike subjective benchmarks such as OwlEval\cite{Ye2023mPLUGOwlME}, which offer comprehensive evaluations of a model’s abilities through human labor but suffer from scalability issues and bias, MMBench offers a more systematic and objective approach.

MMBench consists of two primary components: (i) Curated Dataset. MMBench features a dataset that stands out from existing benchmarks in terms of the number and variety of evaluation questions and abilities (e.g. splits that evaluate emotion or celebrity recognition to test outside knowledge of an LMM). (ii) CircularEval Strategy with ChatGPT Integration. The second key component of MMBench is the CircularEval strategy, which is complemented by the integration of ChatGPT. This approach is aimed at converting free-form predictions into predefined choices, leading to a more robust and reliable evaluation of the LMMs' predictions.

MMBench is thus a comprehensive evaluation pipeline that improves upon existing benchmarks in both scale and depth of assessment.

\minisection{Inference details} We use the official data and code released by the authors which is available at \url{https://github.com/open-compass/MMBench}. All models evaluated on MMBench use the exact method described in Section~\ref{sec:CCoT} of the main paper. Final results are obtained by submitting the output predictions to the official MMBench scoring system at \url{https://mmbench.opencompass.org.cn/mmbench-submission}


\subsection{Winoground}
\minisection{Dataset}
Winoground\cite{winoground} is designed to evaluate the compositional understanding of vision-and-language (VL) models. It challenges these models to correctly pair text and images that share the same underlying compositional structure but differ in the objects involved. Winoground provides a way to assess whether models truly understand the composition of scenes and descriptions, or if they are merely exploiting superficial correlations in the training data.

The Winoground benchmark consists of 400 sets of images and captions. Each set includes two images and two captions, where each image corresponds to one of the captions. The images and captions are carefully designed to be compositionally similar but involve different objects. For example, a set might include an image of a cat chasing a dog with a corresponding caption, and another image of a dog chasing a cat with its caption.

The benchmark evaluates VL models on three scores: (1) Text Score. This score assesses the model's ability to match text captions to the correct images. A high text score indicates that the model effectively understands and applies compositional structures in language.
(2) Image Score. This score evaluates how well the model matches images to the corresponding text captions. A high image score suggests a strong understanding of compositional structures in visual data.
(3) Group Score. This score is the average of the text and image scores. It provides a holistic measure of the model's overall performance in understanding and applying compositional structures across both visual and textual data.

The Winoground benchmark is significant because it moves beyond traditional benchmarks that often allow models to succeed by leveraging simple heuristics or biases in the data. Instead, Winoground requires models to demonstrate a genuine understanding of the compositional relationships between objects in images and their descriptions, demonstrating our model's value in enhancing compositional visual reasoning in LMMs.
\label{supp:impl:vsr}





\minisection{Inference details} We use the official data released by the authors which is available at \url{https://huggingface.co/datasets/facebook/winoground}. Since we evaluate on LMM methods that were designed for single-image inference, we perform a two-step answer extraction process for the image and group tasks (which have two images). (1) First, instead of asking for an answer-choice as in multiple-choice formatted questions, we ask the LMM to generate reasoning for each image caption pair. (2) Secondly, both text reasoning responses are prompted to GPT-4 to yield the LMM's intended answer.

\subsection{WHOOPS!}
\minisection{Dataset} The WHOOPS!\cite{BittonGuetta2023BreakingCSWHOOPS} dataset is a distinctive benchmark developed to evaluate AI models' visual commonsense reasoning, with a particular emphasis on compositional understanding. It consists of 500 synthetic images, each uniquely designed to defy commonsense norms, accompanied by 10874 annotations. These images, crafted using advanced text-to-image models such as Midjourney, DALL-E, and Stable-Diffusion, present scenarios that are logically or physically implausible, thus challenging AI models to go beyond simple object recognition and delve into deeper interpretative reasoning.

The dataset is notable for its diverse array of 'weirdness' categories, encompassing temporal discrepancies, biological rules, cultural knowledge, and more. Each image in WHOOPS! is an invitation for AI models to engage in sophisticated multi-step reasoning, connecting visual cues to knowledge about the world in ways that require a nuanced understanding of everyday experiences, physical and social knowledge, and cultural norms.

WHOOPS! offers four distinct tasks for model evaluation: (i) Explanation Generation. Where models must articulate detailed reasons behind the unusual nature of an image.
Image Captioning: Involving the summarization of the images' content.
Cross-Modal Matching: Requiring models to differentiate between detailed and underspecified captions. (ii) Visual Question Answering (VQA). This task specifically assesses the models’ ability to understand and interpret compositional elements in the images.
In the context of the research paper, the Visual Question Answering (VQA) task of the WHOOPS! dataset was chosen for evaluation. This task is designed to test models' compositional understanding and reasoning. It requires models to answer questions that probe their comprehension of the unusual or 'weird' elements within the images, focusing on their ability to integrate visual information with commonsense knowledge. This task is particularly relevant for assessing how well AI models grasp the implausible or unconventional contexts in which objects are depicted, demanding an advanced level of compositional reasoning. By selecting the VQA task from the WHOOPS! dataset, our work aims to critically evaluate and advance the capabilities of LMM models in compositional visual understanding.

\minisection{Inference details} We use the official data released by the authors which is available at \url{https://whoops-benchmark.github.io/}. For our evaluation on the VQA split, we use the same answer extraction and evaluation process as the paper \cite{BittonGuetta2023BreakingCSWHOOPS}.

\subsection{LLaVA-Bench (In-the-Wild)}
\minisection{Dataset} LLaVA-Bench (In-the-Wild)\cite{liu2023llava15} is a newly developed benchmark that has been used to evaluate the ability of LMMs to provide detailed, yet generalized chat responses to multimodal questions on a variety of images. Given an image, the LMM is prompted with a multimodal task. The LMM's response is compared to GPT-4 generated responses to assess the quality of the response. Although still in development, this small, hand-designed benchmark demonstrates the effectiveness of our method on multimodal chat scenarios.

\minisection{Inference details} We use the official data released by the authors which is available at \url{https://github.com/haotian-liu/LLaVA/blob/main/docs/LLaVA_Bench.md}. For this more open-ended task, we first generate the zero-shot response. Following this, we use our method to generate a scene-graph and then improve the original response. This is to account for the fact that the dataset resembles a long-form conversation.



\section{Qualitative Visualizations}

\label{supp:qual}

We present further qualitative success and failure cases of \textbf{LLaVA-1.5-{\smodel}} in Figure~\ref{fig:supp_examples}.

\section{Licenses and Privacy}
\label{supp:datasets:Licenses}
The license, PII, and consent details of each dataset are in the respective papers. In addition, we wish to emphasize that the datasets we use do not contain any harmful or offensive content, as many other papers in the field also use them. Thus, we do not anticipate a specific negative impact, but, as with any machine learning method, we recommend exercising caution.

\input{figs/fig_4_examples}




%% file: tables/tbl_7_supp.tex
\renewcommand{\arraystretch}{1}
\newcolumntype{?}{!{\vrule width 2 pt}}
\newcolumntype{P}[1]{>{\centering\arraybackslash}p{#1}}

\begin{table*}[htbp]
\begin{center}
\begin{tabular}{m{0.27\textwidth}P{0.05\textwidth}P{0.05\textwidth}P{0.05\textwidth}P{0.05\textwidth}P{0.05\textwidth}P{0.05\textwidth}P{0.05\textwidth} G{0.07\textwidth}}
        \toprule
        Model  & SU & IId & IA & IL & SR & VR &  IIn & W. Avg. \\ \midrule
        \textbf{LLaVA-1.5-13B-{\smodel}} & 76.0 & 74.4 & 71.8 & 64.3 &  54.5& 79.2 & 74.2 & 72.1\\ 
        LLaVA-1.5-13B  & 74.9 & 71.3 &68.9 & 63.5 & 51.5 & 77.0& 73.2&69.9 \\
        \midrule
        \quad w/ Random Scene Graphs & 73.4 & 71.3 & 67.2 &62.2 & 50.2 & 77.3 & 75.3 & 66.5\\ 
        \quad w/out Objects & 76.0 & 73.8 & 71.6 &63.4 & 52.3 & 79.5 & 76.3 & 71.4\\
        \quad w/out Attributes & 75.7 & 73.5 & 71.2 & 63.9 & 52.5 & 79.2 & 72.2 & 70.7\\
        \quad w/out Relationships & 75.4 & 73.1 & 71.8 & 64.3& 52.8 & 79.5 & 74.2 & 71.1\\
        \quad w/out Image & 49.2 & 46.6 & 47.1 & 43.2 & 38.5 & 54.7 & 50.5 & 46.0\\
        \quad w/out "Scene Graph" & 74.7 & 72.3 & 72.5 & 60.4 & 53.0 & 77.0 & 72.2 & 69.8 \\
        \midrule
        \textbf{InstructBLIP-13B-{\smodel}} & 68.7 & 57.9 & 63.7 & 47.9 & 42.8 & 67.1 & 66.0 & 60.1\\
        \quad w/ LLaVA-1.5-13B-{\smodel} SGs & 50.6 & 42.2 & 43.0 & 38.1 & 33.8 & 58.0 & 50.5 & 44.8\\

        \bottomrule
        
    \end{tabular}
        
\end{center}
\caption{\textbf{Ablations on SEEDBench-Image.} This table describes key split-level ablation results of our method on all image splits of SEEDBench~\cite{Li2023SEEDBenchBM}: Instances Counting [IC], Scene Understanding [SU],Instance Identity [IIn], Instance Attributes [IA], Instance Location[IL], Spatial Relation [SR], Visual Reasoning [VR], Text Understanding [TU], Instance Interaction[IIn]. W. Avg. denotes the weighted average.}
\label{tbl: ablations_supp}
\end{table*}

%% file: tables/tbl_3.tex
\renewcommand{\arraystretch}{1}
\newcolumntype{?}{!{\vrule width 2 pt}}
\newcolumntype{P}[1]{>{\centering\arraybackslash}p{#1}}

\begin{table*}[h!]
\begin{center}
\begin{tabular}{m{0.23\textwidth}P{0.05\textwidth}P{0.05\textwidth}P{0.05\textwidth}P{0.05\textwidth}P{0.05\textwidth}P{0.05\textwidth}P{0.05\textwidth}P{0.05\textwidth}P{0.05\textwidth}}
        \toprule
        Model & IC & SU & IId & IA & IL & SR & VR & TU & IIn  \\
        \hline
        InstructBLIP-13B & 29.7 & 60.3 & 55.4 &51.0 & 41.8 & 32.4 & 46.8 & 31.8 & 47.42 \\
        \textbf{InstructBLIP-13B-{\smodel}} & 34.2 & 68.7 & 57.9 & 63.7 & 47.9 & 42.8 & 67.1 & 40.0 & 66.0 \\ \midrule
        LLaVA-1.5-13B & 61.3 & 74.9 & 71.3 &68.9 & 63.5 & 51.5 &  77.04 & 60 &  73.2 \\
        \textbf{LLaVA-1.5-13B-{\smodel}} & 59.3 & 76 & 74.4 & 71.8 & 64.3 & 54.5 & 79.2  & 58.8& 74.2 \\ \bottomrule
    \end{tabular}
    \end{center}
\vspace{-3mm}
\caption{\textbf{Detailed Results Table SEEDBench.} This table describes the split-level results of our method on all image splits of SEEDBench~\cite{Li2023SEEDBenchBM}: Instances Counting [IC], Scene Understanding [SU], Instance Identity [IId], Instance Attributes [IA], Instance Location[IL], Spatial Relation [SR], Visual Reasoning [VR], Text Understanding [TU], Instance Interaction[IIn]]. }
\label{tbl: detailedSEED}
\end{table*}

%% file: tables/tbl_4.tex
\renewcommand{\arraystretch}{1}
\newcolumntype{?}{!{\vrule width 2 pt}}
\newcolumntype{P}[1]{>{\centering\arraybackslash}p{#1}}
\newcolumntype{G}[1]{>{\columncolor{lightgray}\centering\arraybackslash}p{#1}}

\begin{table*}[htbp]
\begin{center}
\begin{tabular}{m{0.23\textwidth}P{0.05\textwidth}P{0.05\textwidth}P{0.05\textwidth}P{0.05\textwidth}P{0.05\textwidth}P{0.05\textwidth}P{0.05\textwidth}P{0.05\textwidth}P{0.05\textwidth}}
        \toprule
        Model & LR & AR & RR & FP-S & FP-C & CP   \\ \hline
        InstructBLIP-13B & 11.5 & 43.6 & 35.5 &36.6 & 22.3 & 51.7   \\ 
        \textbf{InstructBLIP-13B-{\smodel}} & 12.5 & 45.8 & 40.9 & 40.7 & 22.1 & 56.0   \\ \hline
        LLaVA-1.5-13B& 39.9 & 74.7 & 61.6 & 70.9 & 59.9 & 75.4   \\
        \textbf{LLaVA-1.5-13B-{\smodel}} & 44.2 & 72.1 & 75.3 & 73.7 & 59.3 & 81.2  \\ 
        \bottomrule
    \end{tabular}
\end{center}
\vspace{-3mm}
\caption{\textbf{Detailed Results Table MMBench Reasoning.} This table describes the split-level results of our method on splits classified as Reasoning by MMBench~\cite{Liu2023MMBenchIY}: Logic Reasoning [LR], Attribute Reasoning [AR], Relation Reasoning [RR], Fine-Grained(Single) [FG-S], Fine-Grained (Cross) [FG-C], Coarse Perception [CP].}
\label{tbl: MMBReasoning}
\end{table*}

%% file: tables/tbl_6_supp.tex
\renewcommand{\arraystretch}{1}
\newcolumntype{?}{!{\vrule width 2 pt}}
\newcolumntype{P}[1]{>{\centering\arraybackslash}p{#1}}
\newcolumntype{G}[1]{>{\columncolor{lightgray}\centering\arraybackslash}p{#1}}

\begin{table*}[htbp]
\begin{center}
\begin{tabular}{m{0.22\textwidth}P{0.04\textwidth}P{0.04\textwidth}P{0.04\textwidth}P{0.04\textwidth}P{0.04\textwidth}G{0.04\textwidth}G{0.04\textwidth}G{0.04\textwidth}G{0.04\textwidth}P{0.04\textwidth}P{0.04\textwidth}P{0.04\textwidth}}
        &\multicolumn{5}{c}{Coarse Perception}& \multicolumn{4}{c}{FGSI} & \multicolumn{2}{c}{FGCI} \\\hline
        Model & IT & IQ & IE & IS & IS & OCR & CR & OL& ARS & AC & SR   \\ \hline
        InstructBLIP-13B & 16.7 & 14.8 & 50.0 & 37.1 &22.6 & 35.0 & 53.5 & 4.9 &7.1&2.1&1.0 \\ 
        InstructBLIP-13B-ZS-CoT & 16.7 & 3.0 & 36.0 & 36.1 & 20.8 & 37.5 & 39.3& 6.17 &40.4&2.1&2.2\\ 
        \textbf{InstructBLIP-13B-{\smodel}} & 61.1& 9.3 & 54.0 & 82.9& 49.1 & 45.0  & 60.0  & 8.64&45.8 &11.4&4.4\\ \hline
        LLaVA-1.5-13B& 83.3 & 50.0 & 86.0 & 95.2 & 73.6 & 57.5 & 81.8 & 45.7 &87.0&61.4&93.0 \\ 
        LLaVA-1.5-13B-ZS-CoT& 80.5 & 55.6 & 82.0 & 95.2 & 81.1 & 57.5 & 78.8 & 40.7 &92.2&59.1&26.7 \\ 
        \textbf{LLaVA-1.5-13B-{\smodel}} & 81.5 & 44.4 & 86.0 & 97.1 &83.0& 62.5 & 84.8 & 53.1 & 87.0&83.9 &31.1 \\ \hline
    \end{tabular}
    \end{center}
\caption{\textbf{Detailed Results Table MMBench Perception.} This table describes the split-level results of our method on splits classified as Reasoning by MMBench[]. Category Abbreviations:Fine-Grained Perception (Single-Instance) [FGSI], Fine-Grained Perception (Cross-Instance) [FGCI]; Split Abbreviations: Image Topic [IT], Image Quality [IQ], Image Emotion [IE], Image Scene [IS], Image Style [IS], OCR [OCR], Celebrity Recognition [CR], Object Localization [OL], Attribute Recognition (Single-Instance) [ARS], Attribute Recognition (Cross-Instance) [ARC] Attribute Comparison [AC], Spatial Relationship [SR]. }
\label{tbl: MMBenchPerception_supp}
\end{table*}

%% file: figs/fig_4_examples.tex
\begin{figure*}[t]
  \centering
     \includegraphics[width=1.0\linewidth]{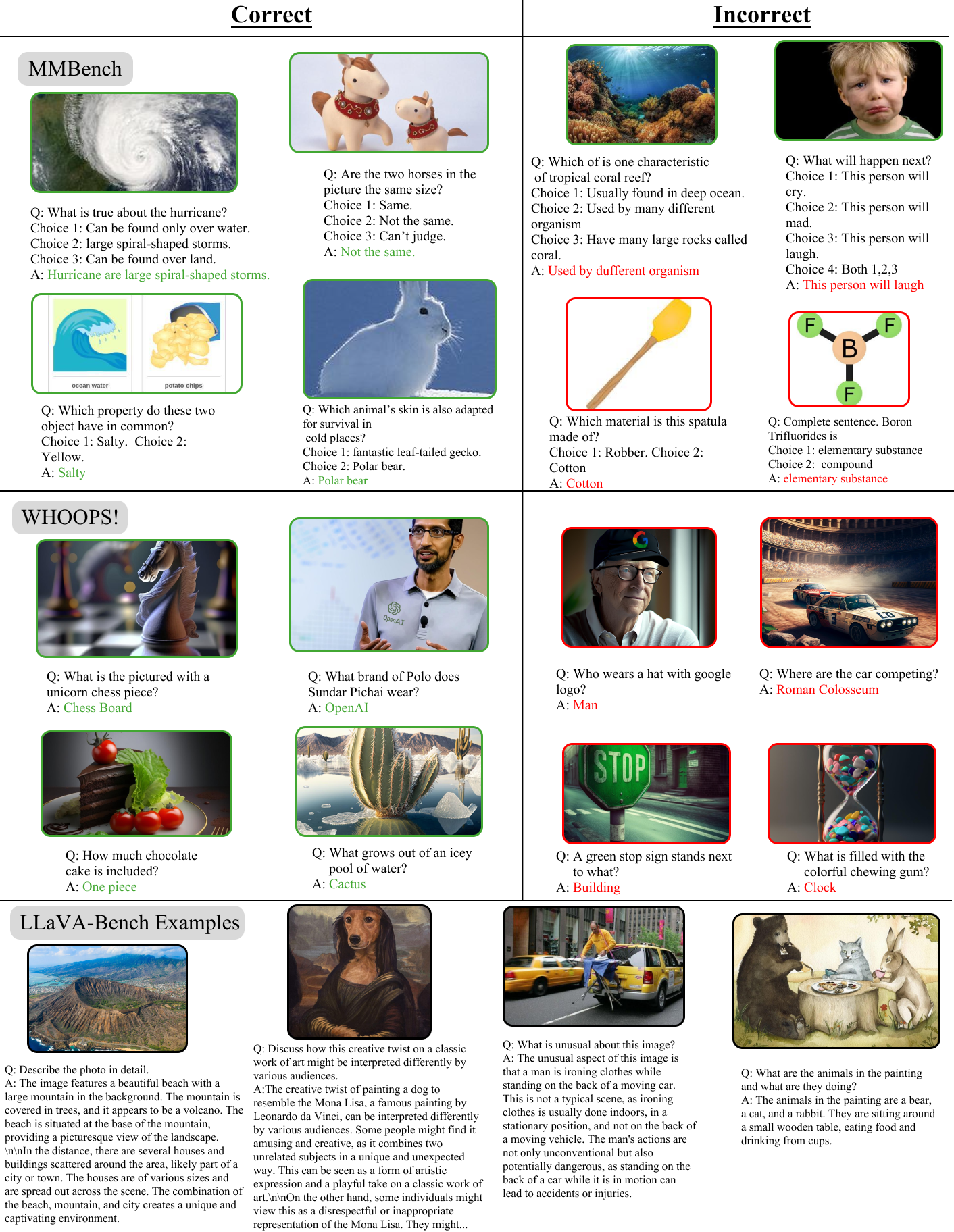}
    \caption{\textbf{Additional Example Outputs.} Above we show some additional examples of our method on both MMBench, WHOOPS!, and LLaVA-Bench.} 
    \label{fig:supp_examples}
\end{figure*}